\documentclass{article}

\usepackage[preprint]{neurips_2019}

\long\def\ignorethis#1{}

\usepackage[utf8]{inputenc} 
\usepackage[T1]{fontenc}    
\usepackage{hyperref}       
\usepackage{url}            
\usepackage{booktabs}       
\usepackage{amsfonts}       
\usepackage{nicefrac}       
\usepackage{microtype}      
\usepackage{graphicx}
\usepackage{wrapfig}
\usepackage{amsmath}
\usepackage{xcolor}
\usepackage{algorithm}
\usepackage{algorithmic, multicol}

\title{Is Fast Adaptation All You Need?}

\author{%
	Khurram Javed, Martha White\\
Department of Computing Science\\
University of Alberta\\
\texttt{kjaved@ualberta.ca, whitem@ualberta.ca} 
\And
	Hengshuai Yao\\
HiSilicon\\
 Huawei Resaerch\\
\texttt{hengshuai.yao@huawei.com} 
}

\begin{document}

\maketitle

\begin{abstract}
	
 Gradient-based meta-learning has proven to be highly effective at learning model initializations, representations, and update rules that allow fast adaptation from a few samples. The core idea behind these approaches is to use fast adaptation and generalization -- two second-order metrics -- as training signals on a meta-training dataset. However, little attention has been given to other possible second-order metrics. In this paper, we investigate a different training signal -- robustness to catastrophic interference -- and demonstrate that representations learned by directing minimizing interference are more conducive to incremental learning than those learned by just maximizing fast adaptation. 
\end{abstract}

\section{Introduction} 

Artificial Neural Networks have proven to be highly successful function approximators when (1) trained on large datasets and (2) trained till convergence using IID sampling. Without large datasets and IID sampling, however, they are prone to over-fitting and catastrophic forgetting \citep{french1991using, french1999catastrophic} respectively. Gradient-based meta-learning has recently been shown to be highly successful at extracting the high-level stationary structure of a problem from a meta-data set -- a dataset of datasets -- allowing few-shot generalization without over-fitting \citep{Finn:EECS-2018-105}. More recently, it has also been shown to mitigate forgetting for better continual learning \citep{nagabandi2018deep, javed2019meta}. 

A gradient-based meta-learner has two important components. (1) The meta-objective -- the objective function that the algorithm minimizes during meta-training -- and meta-parameters -- the parameters updated during meta-training to minimize the selected meta-objective.  One of the most popular realizations of such a meta-learning framework is MAML \citep{finn2017model}. MAML solves few-shot learning by maximizing fast adaptation and generalization as a meta-objective by learning a model initialization -- a set of weights used to initialize the parameters of a neural network. The idea is encode the stationary structure of tasks coming from a fixed task distribution in the weights used for initializing a model such that regular SGD updates starting from this initialization are effective for few-shot learning.

While the choices made by MAML for the meta-objective and meta-parameters are reasonable, there are many other alternatives. For instance, instead of learning a model initialization, we could learn a representation \citep{javed2019meta, bengio2019meta} -- an encoder that transform input data into a vector representation more conducive for learning --, learning rates \citep{li2017meta}, an update rule \citep{bengio1990learning, metz2019meta}, a causal structure \citep{bengio2019meta}, or even the complete learning algorithm \citep{ravi2016optimization}. Similarly, instead of using the few-shot objective, it is possible to define a meta-objective that minimizes other second-order metrics, such as catastrophic forgetting \citep{javed2019meta, riemer2018learning}. 

In this work, we investigate if incorporating robustness to interference in the meta-objective improves performance on incremental learning benchmarks at meta-test time. Recently \citet{javed2019meta} introduced an objective -- \textit{MRCL} -- that learns a representation by minimizing interference and showed that such representations drastically improve performance on incremental learning benchmarks. However, they do not compare their method to representations learned by the few-shot learning objective. \citet{nagabandi2018deep}, on the other hand, found that incorporating effects of incremental learning -- such as interference -- at meta-train time \textit{did not} improve performance on their continual learning benchmark at meta-test time. It is a fair question, then, if the new objective introduced by \citet{javed2019meta} is necessary for effective incremental learning; it is possible that fast adaptation alone would be sufficient for meta-learning non-interfering representations. 

\section{Problem Formulation} 
\label{sec:problem}

\newcommand{\task}{\mathcal{T}}
\newcommand{\loss}{\mathcal{L}}
\newcommand{\inp}{\mathbf{X}}

\newcommand{\out}{\mathbf{Y}}
\newcommand{\learner}{f}
\newcommand{\lossi}{\loss_{\task_i}}

\newcommand{\repparams}{\theta}	
\newcommand{\taskparams}{W}	
\newcommand{\repdim}{d}
\newcommand{\numupdates}{k}
\newcommand{\numy}{m}	
\newcommand{\yhat}{\hat{y}}	
\newcommand{\encoder}{\phi_{\repparams}}

\newcommand{\defeq}{\mathrel{\overset{\makebox[0pt]{\mbox{\normalfont\tiny\sffamily def}}}{=}}}

\newcommand{\CLP}{\text{CLP}}	
\newcommand{\OML}{\text{OML}}	
\newcommand{\MRCL}{\text{MRCL}}

To compare the two objectives, we propose learning Continual Learning Prediction (CLP)  tasks -- a problem setting that requires both fast adaptation and robustness to interference -- online. We define a Continual Learning Prediction (CLP) task as:
\begin{equation*}
 \task = \{  (\inp_1, \out_1), \loss(f(\inp_i), \out_i), q(\inp_{t+1}, \out_{t+1} | \inp_t, \dots, \inp_1), H, \mathcal{X}, \mathcal{Y}\}
 \end{equation*}
  consisting of an initial observation and target ($\inp_1, \out_1$), a loss function $ \loss(f(\inp_i), \out_i)$\footnote{Here $f$ refers to our parametrized model.}   , transition dynamics $q(\inp_{t+1}, \out_{t+1} |  \inp_t, \dots, \inp_1)$, an episode length $H$, and sets $\mathcal{X}, \mathcal{Y}$ such that $\inp_i \in \ \mathcal{X}$ and $\out_i \in \mathcal{Y}$. A sample from a CLP task, $\mathcal S$, consists of a stream of potentially highly correlated samples of length H starting from $\inp_1$ and following the transition dynamics for $H$ steps to get  $\mathcal S = (\inp_1, \out_1), (\inp_2, \out_2), \ldots, (\inp_H, \out_H) \sim \task.$

Furthermore, we define loss over a sample as $\loss(\mathcal S) =\sum_{i=1}^{H} \loss(f(\inp_i), \out_i)$. The learning objective of the CLP task is to minimize the expected loss of a task i.e. $\mathbb{E_{\mathcal S \sim \task}}[ \loss(S)]$ from a single sample $\mathcal S_{train}$ by seeing one data point at a time. Standard neural network, without any meta-learning, applied to the CLP task would do poorly as they struggle to learn online from a highly correlated stream of data in a singe pass. 

\section{Comparing the Two Objectives} 

\begin{wrapfigure}{R}{0.55\textwidth}
	\vspace{-20pt}
	\begin{minipage}{0.55\textwidth}
		\begin{algorithm}[H]
			\centering
			\caption{Meta-Training : MAML Objective}\label{MAMLalgorithm}
			\begin{algorithmic}[1]
				\REQUIRE $p(\task)$: distribution over tasks
				\REQUIRE $\alpha$, $\beta$: step size hyperparameters
					\REQUIRE $K$: No of inner gradient steps
				\STATE randomly initialize $\theta$
				\WHILE{not done}
				\STATE Sample task $\task_i \sim p(\task)$
				
				\STATE Sample $\mathcal S_{train}^i$ from $\task_i$
				\STATE $\taskparams_0 = \taskparams$
				\color{red} \FOR {$j$ in $1, 2, \dots , K$}
				\STATE   \color{red} $\taskparams_j=\taskparams_{j-1}-\alpha
				\nabla_{\taskparams_{j-1}}  \lossi(\mathcal S_{train}^i, f_{\repparams, \taskparams_{j-1}})$ 
				\ENDFOR
				\color{black}
				\STATE Sample $S_{test}^i$ from $\task_i$
				
				\STATE Update $\repparams \leftarrow \repparams - \beta \nabla_\repparams \lossi (S_{test}^i, f_{\repparams, \taskparams_K})$ 
				\ENDWHILE
			\end{algorithmic}
		\end{algorithm}
	\end{minipage}
	\vspace{-10pt}
\end{wrapfigure}

To apply neural network to the CLP task, we propose meta-learning a function $\phi_{\repparams}(\inp)$ -- a deep neural network parametrized by $\repparams$ -- from $\mathcal{X}  \rightarrow  \mathbb{R}^\repdim$. We then learn another function $g_{\taskparams}$ from $\mathbb{R}^\repdim \rightarrow  \mathcal{Y}$. By composing the two functions we get $f_{\taskparams, \repparams}(\inp) = g_\taskparams(\phi_\repparams(\inp))$ which constitute our model for the CLP tasks.  We treat $\repparams$ as meta-parameters that are learned by minimizing the meta-objective and then later fixed at meta-test time. After learning $\repparams$, we learn $g_{\taskparams}$ from $\mathbb{R}^\repdim \rightarrow  \mathcal{Y}$ for a CLP task from a single trajectory $\mathcal S_{train}$ using fully online SGD updates in a single pass.  

 For meta-training, we assume a distribution over CLP tasks given by $p(\task)$. We consider two meta-objectives for updating the meta-parameters $\repparams$. (1) A MAML like few-shot-learning objective, and MRCL -- an objective that also minimizes interference in addition to maximizing fast adaptation. The two objectives can be implemented as Algorithm~\ref{MAMLalgorithm} and \ref{algorithm} respectively with the primary difference between the two highlighted in red. Note that MAML uses the complete batch of data $\mathcal S_{train}$ to do $K$ inner updates where MRCL uses one data point from $\mathcal S_{train}$ for one update. This allows MRCL to take the effects of incremental learning -- such as catastrophic forgetting -- into account.

\newcommand{\target}{\mathbf{y}}

\section{Dataset, Implementation Details, and Results}

\subsection{CLP tasks using Omniglot}

 Omniglot is a dataset of over 1623 characters from 50 different alphabets \citep{lake2015human}. Each character has 20 hand-written images. The dataset is divided into two parts. The first 963 classes constitute the meta-training dataset whereas the remaining 660 the meta-testing dataset. To define a CLP task on these datasets, we sample an ordered set of 200 classes $(C_{1},C_{2},C_{3}, \dots, C_{200})$. $\mathcal X$ and $\mathcal Y$, then, constitute of all images of these classes. A sample $\mathcal S$ from such a task is a trajectory of images -- five images per class -- where we see all five images of $C_1$ followed by five images of $C_2$ and so on. This makes $H= 5 \times 200 = 1000$. Note that the sampling operation defines a distribution $p(\task)$ over tasks which we use for meta-training.

\begin{wrapfigure}{R}{0.55\textwidth}
	\vspace{-25pt}
	\begin{minipage}{0.55\textwidth}
		\begin{algorithm}[H]
			\centering
			\caption{Meta-Training : MRCL Objective}\label{algorithm}
			\begin{algorithmic}[1]
				\REQUIRE $p(\task)$: distribution over tasks
				\REQUIRE $\alpha$, $\beta$: step size hyperparameters
				\STATE randomly initialize $\repparams, \taskparams$
				\WHILE{not done}
				\STATE Sample task $\task_i \sim p(\task)$
				\STATE Sample $\mathcal S_{train}^i$ from $\task_i$
				\STATE $\taskparams_0 = \taskparams$
				\color{red}
				\FOR {$j = 1, 2, \dots, |\mathcal S_{train}^i$|}
				\STATE  $\taskparams_j=\taskparams_{j-1}-\alpha
				\nabla_{\taskparams_{j-1}}  \lossi(\inp_{j}^i, f_{\repparams, \taskparams_{j-1}})$ 
				\ENDFOR
				\color{black}
				\STATE Sample $S_{test}^i$ from $\task_i$
				
				\STATE Update $\repparams \leftarrow \repparams - \beta \nabla_\repparams \lossi (S_{test}^i, f_{\repparams, \taskparams_{|\mathcal S_{train}|}})$ 
				\ENDWHILE
			\end{algorithmic}
		\end{algorithm}
	\end{minipage}
	\vspace{-10pt}

\end{wrapfigure}

\subsection{Meta-Training}
We learn an encoder -- a deep CNN with 6 convolution and two FC layers -- using the MAML and the MRCL objective. We treat the convolution parameters as $\repparams$ and FC layer parameters as $\taskparams$. Because optimizing the MRCL objective is computationally expensive for $H=1000$ (It involves unrolling the computation graph for 1,000 steps), we approximate the two objectives. For MAML we learn the $\phi_\repparams$ by maximizing fast adaptation for a 5 shot 5-way classifier. For MRCL, instead of doing $|\mathcal S_{train}|$ no of inner-gradient steps as described in Algorithm~\ref{algorithm}, we go over $\mathcal S_{train}$ five steps at a time. For $kth$ five steps in the inner loop, we accumulate our meta-loss on $\mathcal S_{test}[0:5\times k]$, and update our meta-parameters using these accumulated gradients at the end as explained in Algorithm \ref{algorithm_approx} in the Appendix.  This allows us to never unroll our computation graphs for more than five steps (Similar to truncated back-propagation through time) and still take into account the effects of interference at meta-training.

Finally, both MAML and MRCL use 5 inner gradient steps and similar network architectures for a fair comparison. Moreover, for both methods, we try multiple values for the inner learning rate $\alpha$ and report the results for the best parameter. For more details about hyper-parameters see the Appendix.

 \begin{wrapfigure}{R}{0.55\textwidth}

 		\vspace{-30pt}
 	\begin{minipage}{0.55\textwidth}

 		\begin{algorithm}[H]

 			\centering

 			\caption{Meta-Testing}\label{algorithm_test}

 			\begin{algorithmic}[1]

 				\REQUIRE $\task$: 
Given CLP task
 				\REQUIRE $\alpha$: step size hyperparameters

 				\REQUIRE $\repparams^*$: Meta-learned encoder parameters 
 				\STATE randomly initialize $\taskparams$

 				\STATE Sample $\mathcal S_{train}$ from $\task$
 				\STATE $\taskparams_0 = \taskparams$
 				\FOR {$j = 1, 2, \dots, |\mathcal S_{train}$|}
 				\STATE  $\taskparams_j=\taskparams_{j-1}-\alpha
 				\nabla_{\taskparams_{j-1}}  \loss_{\task}(\inp_{j}, f_{\repparams, \taskparams_{j-1}})$ 
 				\ENDFOR
 				\STATE Compute train error (or accuracy) $\loss_{\task}(\mathcal S_{train})$. 
 				\STATE Approximate test error (or accuracy) $\mathbb{E_{\mathcal S \sim \task}}[ \loss_{\task}(S)]$ using multiple samples.

 			\end{algorithmic}

 		\end{algorithm}

 	\end{minipage}

 	\vspace{-30pt}

 \end{wrapfigure}

\textbf{SR-NN}
 \citep{liu2019utility} does not use gradient-based meta-learning; instead, it uses the meta-training dataset to learn a sparse representation by regularizing the activations in the representation layer and serves as a baseline.

\subsection{Meta-Testing}
	\begin{figure}
	\centering
	
	\includegraphics[width=\textwidth]{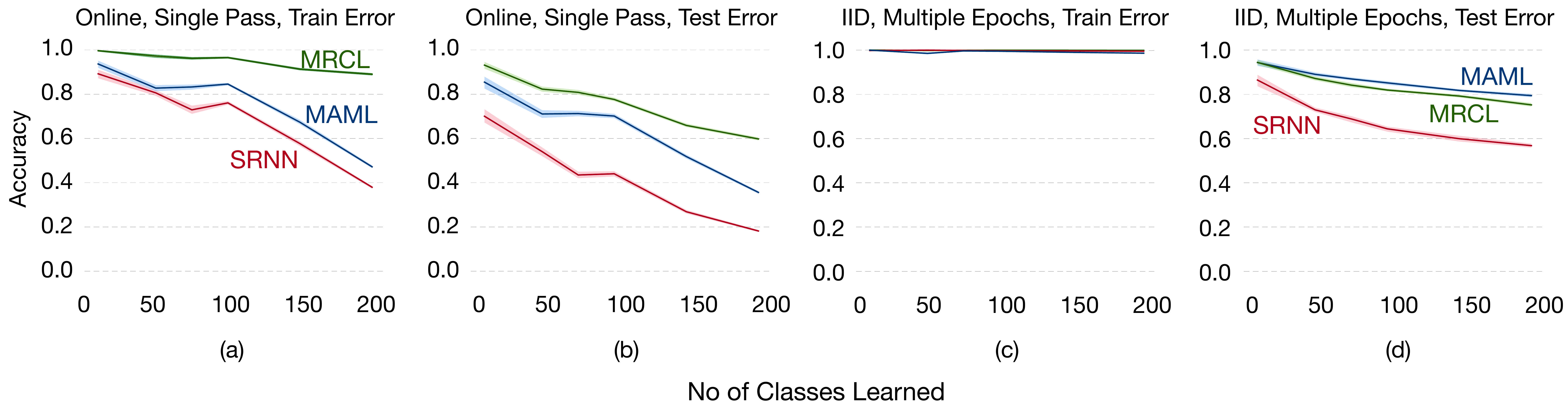}
	
	\caption{Comparison of representations learned by the MAML and MRCL objective for incremental learning. All curves are averaged over 50 CLP tasks with 95\% confidence intervals. At every point on the x-axis, we only report accuracy on the classes seen so far. Even though both MRCL and MAML learn representations that result in comparable performance of classifiers trained under the IID setting (c and d),  MRCL out-performs MAML when learning online on a highly correlated stream of data.}
	\label{fig:imagenet}
\end{figure}

At meta-test time, we sample 50 CLP tasks from the meta-test-set. For each task, we learn $\taskparams$ from a single trajectory $S_{train}$ using Algorithm~\ref{algorithm_test} and compute accuracy on $\mathcal S_{train}$ (Train accuracy). We also measure accuracy on multiple other samples from the task and report them as test accuracy. 

More concretely, we transform all the images in a task to a vector representation $\mathbb{R}^\repdim$ using our meta-learned encoder $\phi_{\repparams}$ and learn a classifier (Up to 200 classes) parametrized by $\taskparams$ fully online (Seeing all the data of one class before moving to the next) in a single pass. We report the accuracy in Fig. \ref{fig:imagenet} (a) and (b) respectively. At every point on the x-axis, we only report accuracy for the classes seen so far (This is why accuracy drops for all methods as we learn more and more classes). We can see from Fig~\ref{fig:imagenet} (a) that representations learned by MRCL are significantly more robust to catastrophic interference than those learned by MAML. Moreover, from Fig~\ref{fig:imagenet} (b), we see that that the higher training accuracy also results in better generalization performance (i.e. MRCL is not just memorizing the training samples).

As a sanity check, we also trained classifiers by sampling data IID for three epochs and report the results in Fig. \ref{fig:imagenet} (c) and (d). The fact that MAML and MRCL do equally well with IID sampling indicates that the quality of representations ($\phi_{\repparams} = \mathbb{R}^\repdim$) learned by both objectives are comparable and the higher performance of MRCL is indeed because the representations are more suitable for incremental learning. 
\section{Discussion}  
\textbf{Intuition Behind the Difference Between MRCL and MAML:}

At an intuitive level, the primary difference between MRCL and MAML is in the inner gradient steps. For MAML, the inner gradient consists of SGD updates on a batch of data from all the classes. As a result, the objective is only maximizing fast adaptation and generalization. For MRCL, on the other hand, the inner gradient steps involve online SGD updates on a highly correlated stream of data. Consequentially, the model not only has to adapt to the task from a single trajectory but it also has to prevent subsequent inner updates from interfering with the earlier updates.  This motivates the model to learn a representation that prevents forgetting of past knowledge. 

\textbf{Why Learn an Encoder as Opposed to a Network Initialization?}

In this work, we meta-learned a representation given by $\encoder$ as opposed to a network initialization. We empirically found that for online learning on highly correlated data-streams, a network initialization is an ineffective inductive bias. This is especially true when learning long trajectories involving thousands of SGD updates. For a more detailed explanation with some empirical results, see Fig. \ref{RLN} in the appendix. 
\section{Conclusion}
In this paper, we compared two meta-learning objectives for learning representations conducive for incremental learning. We found that MRCL -- an objective that directly minimizes interference -- is significantly better at learning such representations than MAML -- an objective that only maximizes generalization and fast adaptation. This is contrary to what \citet{nagabandi2018deep} found in their work. One explanation of why they didn't see the benefit of incorporating online learning in meta-training is that, in their work, they also have a mechanism for detecting changes in tasks. Based on the detected task, an agent might choose to use a different neural network as model. Such a task selection mechanism may make reducing interference less important. This is further supported by looking at \textit{continued adaptation with meta-learning} -- one of the baselines in their paper that uses a single model for continuous adaptation. For this baseline, they did observe that an initialization learned by optimizing the MAML objective was ineffective at preventing forgetting.

\clearpage
\bibliographystyle{named}
\bibliography{noniid}
\clearpage

\appendix

		\begin{algorithm}[H]
			\centering
			\caption{Meta-Training : Approximate Implementation of the MRCL Objective}\label{algorithm_approx}
			\begin{algorithmic}[1]
				\REQUIRE $p(\task)$: distribution over tasks
				\REQUIRE $\alpha$, $\beta$: step size hyperparameters
				\REQUIRE $m$: No of inner gradient steps per update before truncation
				\STATE randomly initialize $\repparams, \taskparams$
				\WHILE{not done}
				\STATE Sample task $\task_i \sim p(\task)$
				\STATE Sample $\mathcal S_{train}^i$ from $\task_i$
				\STATE $\taskparams_0 = \taskparams$
				\STATE $\nabla_{accum} = \mathbf{0}$
				\WHILE {$j \le |\mathcal S_{train}$|}
				\FOR{$k$ in $1, 2, \dots , m$}
				\STATE  $\taskparams_j=\taskparams_{j-1}-\alpha
				\nabla_{\taskparams_{j-1}}  \lossi(\inp_{j}^i, f_{\repparams, \taskparams_{j-1}})$ 
				\STATE $j$ = $j + 1$
				\ENDFOR
				\STATE Sample $S_{test}^i$ from $\task_i$
				\STATE $\nabla_{accum} = \nabla_{accum} +   \nabla_\repparams \lossi (S_{test}^i[0:j], f_{\repparams, \taskparams_{j}})$
				\STATE Stop Gradients$(f_{\repparams, \taskparams_{j}}))$
				\ENDWHILE
				\STATE Update $\repparams \leftarrow \repparams - \beta \nabla_{accum} $ 
				\ENDWHILE
			\end{algorithmic}
		\end{algorithm}

 \begin{wrapfigure}{R}{0.50\textwidth}

	\begin{center}
		\includegraphics[width=0.50\textwidth]{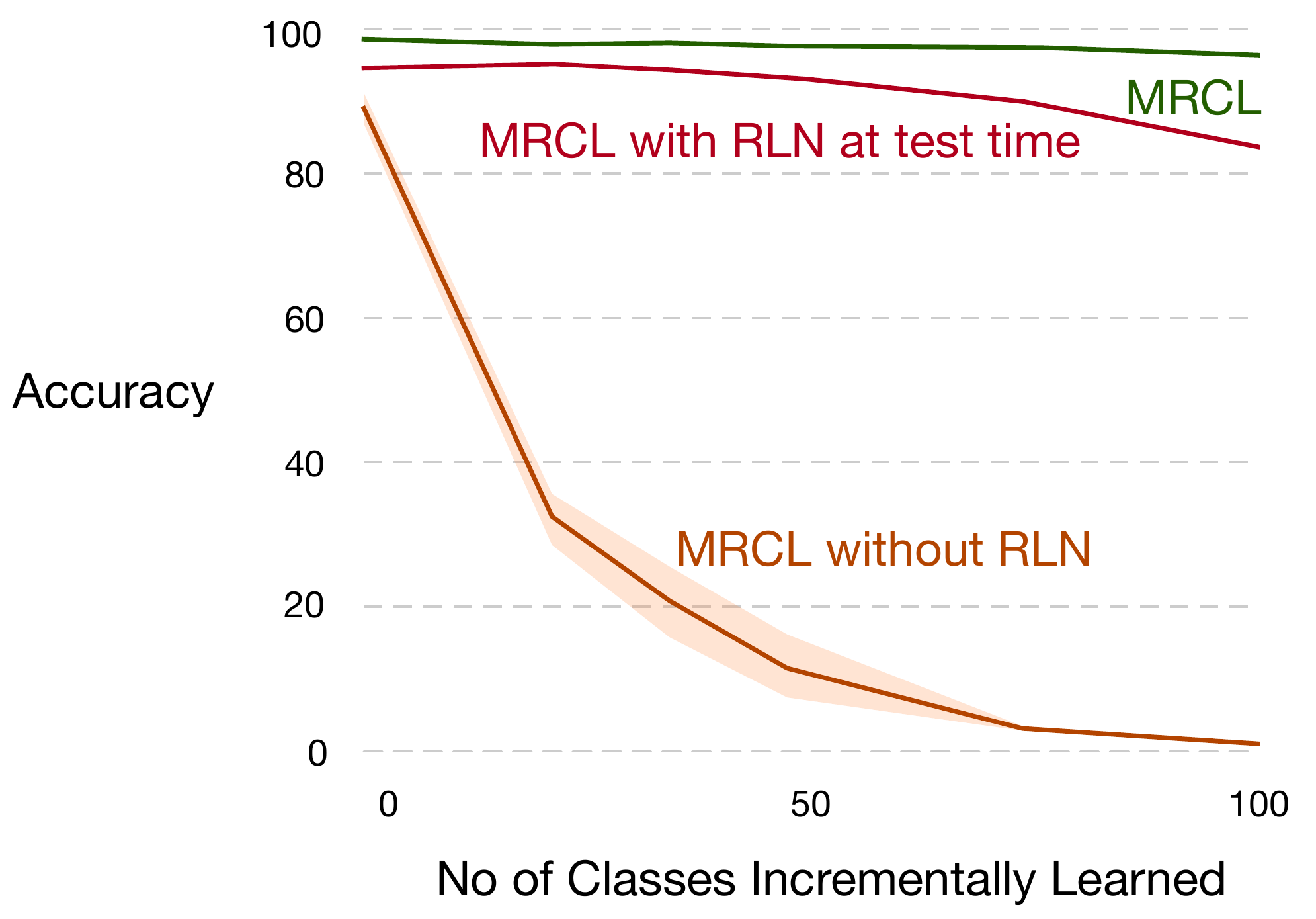}
	\end{center}
	\caption{Instead of learning an encoder $\encoder$ , we learn an initialization by updating both $\theta$ and $W$ in the inner loop of meta-training. In "MRCL without RLN," we also update both at meta-test time whereas in "MRCL without RLN at test time," we fix $theta$ at meta-test time just like we do for MRCL. For each of the methods, we report the training error during meta-testing. It's clear from the results that a model initialization is not an effective bias for incremental learning. Interestingly, "MRCL with RLN at test time" doesn't do very poorly. However, if we know we'll be fixing $\theta$ at meta-test time, it doesn't make sense to update it in the inner loop of meta-training (Since we'd want the inner loop setting to be as similar to meta-test setting as possible.}
	\label{RLN}
\end{wrapfigure}

\subsection{Why Learn an Encoder Instead of Initialization : Explanation}
We empirically found that learning an encoder results in significantly better performance than learning just an initialization as shown in Fig \ref{RLN}. Moreover, the meta-learning optimization problem is more well-behaved when learning an encoder (Less sensitive to hyper-parameters and converges faster). One explanation for this difference is that a global and greedy update algorithm -- such as gradient descent -- will greedily change the weights of the initial layers of the neural network with respect to current samples when learning on a highly correlated stream of data. Such changes in the initial layers will interfere with the past knowledge of the model. As a consequence, an initialization is not an effective inductive bias for incremental learning. When learning an encoder $\encoder$, on the other hand, it is possible for the neural network to learn highly sparse representations which make the update less global (Since weights connecting to features that are zero remain unchanged).  

\begin{table}
	\caption{Hyper-Parameters for Omniglot Representation Learning for MRCL and MAML. Inner learning rate is the only sensitive parameter for both methods. We tried 12 Inner learning rate in the range 1.0 to 1-e6 and picked the best for each method. For MAML, we report results using Inner-LR of 0.5 whereas for MRCL, 0.03.}
	\label{hyper_omni}
	\centering
	\begin{tabular}{lll}
		\toprule
		Parameter     & Description     & Value \\
		\midrule
		Meta LR & Learning rate used for the meta-update  & 1e-4     \\
		Meta update optimizer     & Optimizer used for the meta-update & Adam      \\
		Inner LR     & LR used for the inner updates (MAML, MRCL) & 0.5, 0.03     \\
		Inner LR Search    & Inner LRs tried before picking the best & [1.0, 1e-6]  \\
		Inner steps    & Number of inner gradient steps    & 5  \\
		Conv-layers     & Total convolutional layers    &  6  \\
		FC Layers    & Total fully connected layers    &  2  \\
		Encoder    & Layers in $\phi_\repparams$    &  6  \\
		Kernel     & Size of the convolutional kernel   &  3x3 \\
		Non-linearly     & Non-linearly used   & relu \\
		Stride    & Stride for convolution operation in each layer    &  [2,1,2,1,2,2]  \\
		\# kernels		     & Number of convolution kernels in each layer    &  256 each \\
		Input    & Dimension of the input image   &  84 x 84   \\
		\bottomrule
	\end{tabular}
\end{table}

\end{document}